\documentclass[10pt,letterpaper]{article}
\usepackage{booktabs}
\usepackage{xurl}

\usepackage{threeparttable}
\usepackage{threeparttablex}
\usepackage{multirow}
\usepackage{graphicx}

\usepackage{tabularx}
\usepackage[top=0.85in,left=2.75in,footskip=0.75in]{geometry}

\usepackage{amsmath,amssymb}

\usepackage{changepage}
\usepackage[T1]{fontenc}
\usepackage[utf8]{inputenc}

\usepackage{textcomp,marvosym}
\usepackage{dblfloatfix}
\usepackage{titlesec}
\usepackage{cite}

\usepackage{nameref}
\usepackage[colorlinks=true, linkcolor=blue, citecolor=blue, urlcolor=blue]{hyperref}
\usepackage{caption}
\usepackage[utf8]{inputenc}
\usepackage{makecell}
\usepackage[final]{microtype}

\ifPDFTeX
\DisableLigatures[f]{encoding = *, family = * }
\fi

\usepackage[table]{xcolor}

\usepackage{array}

\newcolumntype{+}{!{\vrule width 2pt}}

\newlength\savedwidth

\raggedright
\usepackage[aboveskip=1pt,labelfont=bf,labelsep=period,justification=raggedright,singlelinecheck=off]{caption}

\makeatletter
\renewcommand{\@biblabel}[1]{\quad#1.}
\makeatother

\usepackage{lastpage,fancyhdr,graphicx}
\usepackage{epstopdf}
\usepackage{float}
\fancyheadoffset[L]{2.25in}
\fancyfootoffset[L]{2.25in}
\begin{document}
\vspace*{0.2in}

\begin{flushleft}
{\Large
\textbf{Estimating Grammatical Gender Directions in Contextual Embeddings under Controlled and Natural Contexts} 
}
\newline
\\
Huanping Xiao\textsuperscript{1},
Yingji Li\textsuperscript{2*}
\\
\bigskip
\textbf{1} College of Foreign Languages and Cultures, Jilin University, Changchun, Jilin Province, China
\\
\textbf{2} College of Computer Science and Technology, Jilin University, Changchun, Jilin Province, China
\bigskip

%
%





* Corresponding author: yingjili@jlu.edu.cn

\end{flushleft}


%
\section*{Abstract}

Contextual language models conflate grammatical gender and social semantic bias in gendered languages such as Spanish. Existing gender debiasing approaches only operate on static word embeddings leaving contextual representations unexplored for this two dimensional gender disentanglement. To address the this issue, we make the first attempt to disentangle grammatical gender from semantic contamination for contextual embeddings. We construct both controlled templates and natural Wikipedia contexts to build balanced datasets of inanimate nouns, and design a framework equipped with centroid, Support Vector Machine (SVM) and Linear Discriminant Analysis (LDA) gender direction estimators as well as contamination-aware weighting strategies. A set of dual-objective evaluation metrics is proposed to balance the suppression of grammatical gender leakage on inanimate nouns and the preservation of semantic gender distinctions for occupation terms. The results reveal that unweighted controlled contexts yield the purest grammatical gender direction, and the centroid estimator achieves better performance than discriminative baselines.

\clearpage
\newgeometry{top=0.85in,left=1in,right=1in,footskip=0.75in}


\section{Introduction}

Pre-trained contextual language models have become foundational tools for multilingual natural language processing~\cite{gamboa-etal-2025-social}. Extensive research~\cite{bolukbasi2016man,kurita-etal-2019-measuring, caliskan2017semantics} has confirmed that these models encode pervasive social gender stereotypes within their word and sentence representations, motivating a wide range of embedding debiasing techniques based on orthogonal projection~\cite{bolukbasi2016man, zhou-etal-2019-examining,dev-etal-2021-oscar}. Most existing bias mitigation approaches are designed for English~\cite{gallegos2024bias,li2023survey,liu-etal-2020-multilingual-denoising}, a language without grammatical gender, and treat all gender-related embedding signals as unwanted social bias~\cite{zhou-etal-2019-examining}. This design creates critical limitations when applied to grammatically gendered languages~\cite{garrido-munoz-etal-2023-maria} like Spanish and French, where two distinct gender signals coexist: mandatory morphosyntactic grammatical gender~\cite{Corbett_1991} and socially constructed semantic gender bias.

Grammatical gender is an inherent linguistic property of inanimate nouns in gendered languages. Nouns are fixed as masculine or feminine and require consistent agreement with articles, adjectives and pronouns. Distributional co-occurrence patterns between nouns and their modifiers make grammatical gender cues automatically detectable in contextual embeddings. Unfortunately, grammatical gender signals are inherently entangled with social semantic gender associations learned from real-world corpora~\cite{bordia-bowman-2019-identifying}. Prior psycholinguistic~\cite{zunino-etal-2025-dresses,boroditsky-etal-2003-sex} and computational~\cite{mihaylov-shtedritski-2024-elegant-bridge, stanczak-etal-2024-causal,sukumaran-etal-2024-investigating,saeed-etal-2025-beyond} studies further prove that even semantically neutral inanimate nouns absorb contextual gendered descriptions from natural text, leading to severe semantic contamination.

While a small body of literatures~\cite{zhou-etal-2019-examining, sabbaghi-caliskan-2022-measuring} have proposed disentanglement pipelines to separate the two gender dimensions, all these existing methods~\cite{zhou-etal-2019-examining, sabbaghi-caliskan-2022-measuring,mccurdy-serbetci-2020-grammatical} rely exclusively on static word embeddings trained from unconstrained natural corpus contexts, and no prior work has attempted such disentanglement within contextual pre-trained language models.

To address these research gaps, we systematically investigate how to construct the purest grammatical gender directions within contextual word embeddings, and further establish a unified analytical framework for disentangling grammatical gender signals from semantic gender bias. We first build balanced datasets of Spanish and French inanimate nouns, paired with two types of context resources: controlled templates for grammatical signal extraction, and natural sentences for contamination quantification. To generate robust grammatical gender axes, we implement centroid-based~\cite{bolukbasi-etal-2016-man}, SVM-based~\cite{bollmann2019forgetting} and LDA-based~\cite{zhou-etal-2019-examining} estimators, and further introduce mean-normalized and median-normalized weighting strategies to down-weight semantically contaminated nouns. We design three complementary evaluation metrics to balance two conflicting goals: suppressing redundant grammatical gender leakage on inanimate nouns while preserving semantic gender distinctions for occupation vocabulary. We adopt the harmonic trade-off score as our primary comprehensive metric to compare all model variants. Experiments on four monolingual and multilingual pre-trained encoders yield consistent conclusions: controlled unweighted contexts produce the purest grammatical gender directions, and the simple centroid estimator outperforms discriminative alternatives. Our orthogonal projection only removes components aligned with grammatical gender axes without disrupting general lexical semantic structure, as validated by supplementary semantic stability tests.

The main contributions of this paper are summarized as follows:

\begin{enumerate}
\item We pioneer grammatical-semantic gender disentanglement for contextual language models, while all prior comparable work is limited to static word embeddings. We build paired controlled template and natural Wikipedia datasets for Spanish/French inanimate nouns to verify that controlled templates eliminate corpus noise and extract cleaner grammatical gender signals.
\item We are the first to formalize semantic contamination in inanimate noun representations for contextual language models and design contamination-aware strategies to suppress biased distributional artifacts.
\item Drawing on WEAT~\cite{caliskan2017semantics}, SEAT~\cite{may-etal-2019-measuring} and hard debiasing~\cite{bolukbasi-etal-2016-man}, we propose a novel dual-objective evaluation framework with leakage reduction, semantic preservation and harmonic trade-off metrics to resolve the lack of balanced evaluation standards for gendered language debiasing.
\end{enumerate}

\begin{figure}[!t]
\centering
\includegraphics[width=\linewidth,height=0.85\textheight,keepaspectratio]{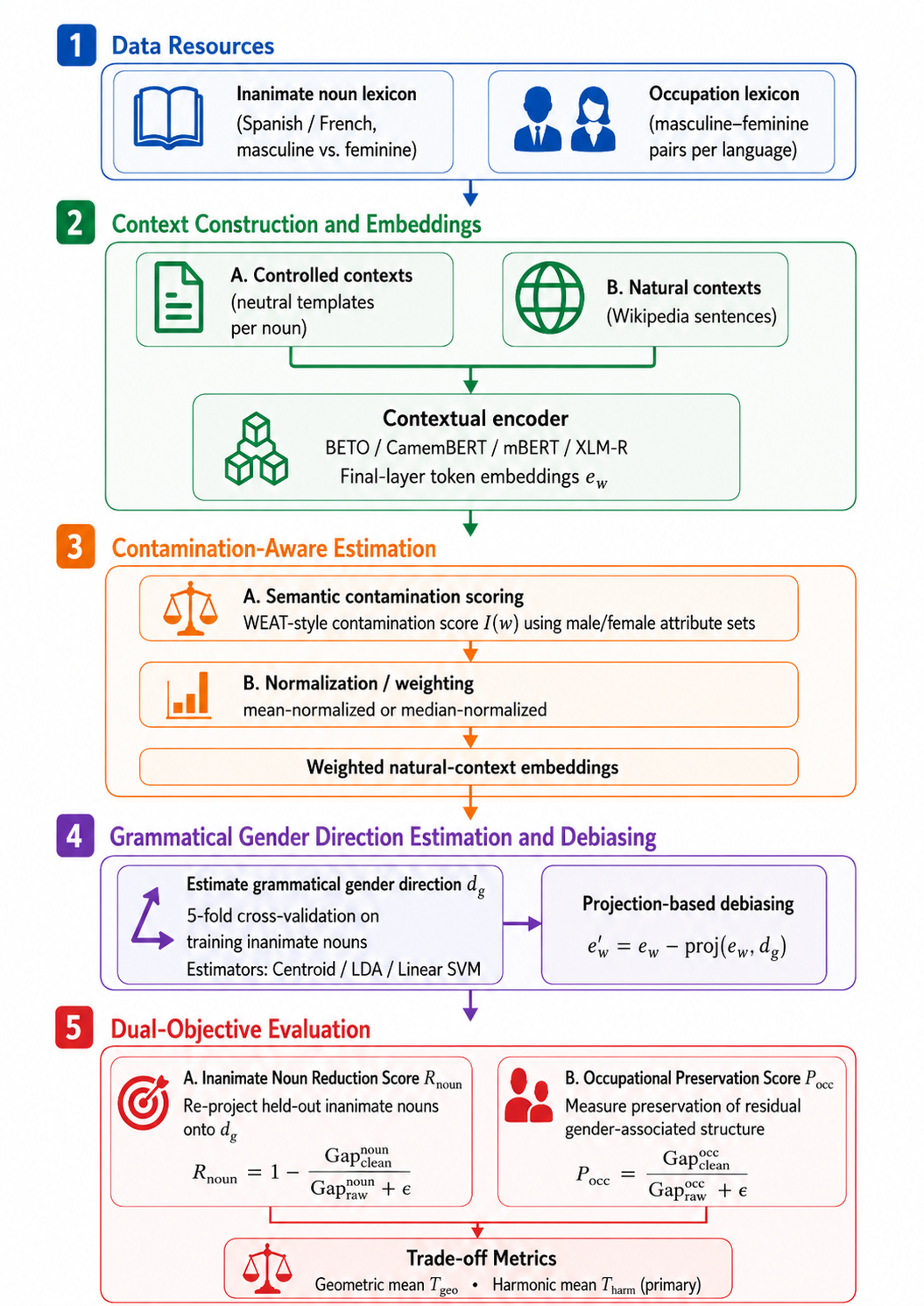}
\caption{\textbf{Overall framework of our grammatical-semantic gender disentanglement method.}}
\label{fig:framework}
\end{figure}

In the following sections, we first review related work on embedding bias, gender signal disentanglement and multilingual bias evaluation. Next, we detail our datasets, context construction and pre-trained models. We then formalize our method including gender direction estimation, contamination weighting and evaluation metrics. After presenting experimental results and ablation analyses, we discuss the core findings, limitations and future research directions. ~\hyperref[fig:framework]{Fig~\ref{fig:framework}} visualizes our unified disentanglement framework.

\section{Related Work}

\subsection{Gender Bias in Embeddings}
Early works\cite{bolukbasi-etal-2016-man, gonen-goldberg-2019-lipstick} firstly revealed that static word embedding techniques such as Word2Vec and GloVe implicitly encode human-like social gender stereotypes within dense vector representations. These classical studies observed that occupation-related words like \textit{engineer}, \textit{doctor} were geometrically closer to masculine pronouns in the embedding space, while words such as \textit{nurse}, \textit{secretary} were tightly clustered with feminine lexical items~\cite{dearteaga-etal-2019-bias}, which replicated the long-standing gender prejudice inherited from training corpora. To mitigate such unwanted bias, researchers pioneered projection-based debiasing pipelines~\cite{dev-etal-2021-oscar}: these methods first identify a low-dimensional gender subspace via principal component analysis from gender-pair word vectors, then project all word embeddings onto the orthogonal complement of this bias subspace to erase gender directional information while striving to preserve the original semantic utility of word vectors. Despite their wide adoption, these early static debiasing strategies suffer from inherent limitations: they apply a globally fixed bias elimination transformation to every token, failing to distinguish between contextually essential gender attributes (e.g., gender information for nouns referring to specific individuals) and harmful stereotypical bias, which inevitably leads to over-debiasing and sacrifices downstream task performance to a certain extent.

With the prevalence of pre-trained contextual language models, later studies\cite{kurita-etal-2019-measuring, may-etal-2019-measuring,  yarrabelly-etal-2024-mitigating} extended gender bias investigation from static word embeddings to contextual token embeddings and sentence-level semantic representations. Different from fixed vector outputs of static embedding methods, BERT-style models generate dynamic token representations conditioned on surrounding textual context, which introduces more complicated bias distribution patterns. Prior studies~\cite{tan-celis-2019-assessing, de-vassimon-manela-etal-2021-stereotype} have found evidence that gender bias persists in transformer-based contextual encoders such as BERT, while contextualized bias measurements can vary substantially depending on sentence context and evaluation method. 
\subsection{Disentanglement of Grammatical and Semantic Gender in Word Representations}
Gendered languages require a strict distinction between grammatical gender and semantic gender, as these two types of gender signals are inherently entangled within lexical representations. Zhou et al.~\cite{zhou-etal-2019-examining} pioneered a disentanglement paradigm that separately estimates grammatical and semantic gender directions in embedding space, and further eliminates grammatical gender components from semantic gender vectors to isolate social gender bias.
A body of follow-up studies~\cite{gonen-etal-2019-grammatical, stanczak-etal-2024-causal} further verifies that grammatical gender information is implicitly encoded in word embeddings, which originates from the fixed co-occurrence patterns between inanimate nouns and their modifiers such as articles and adjectives. Such distributional regularities make it inevitable for grammatical gender to interfere with the measurement of semantic gender bias in gendered languages~\cite{fridriksdottir-einarsson-2024-gendered}.
\subsection{Semantic Bias Embodied in Inanimate Nouns}
Psycholinguistic works~\cite{boroditsky-etal-2003-sex, elpers-etal-2022-grammatical} have debated whether grammatical gender shapes human cognitive perceptions and descriptive tendencies toward inanimate objects.
Recent LLM-based computational study~\cite{mihaylov-shtedritski-2024-elegant-bridge} revisits this classic experimental question from a corpus-driven perspective: researchers prompt multilingual large language models to generate descriptive adjectives for grammatically gendered nouns, then train classification models to predict the grammatical gender of nouns solely from these descriptive texts. The high prediction accuracy achieved across experimental settings strongly confirms that inanimate nouns are contaminated by topical, lexical, and co-occurrence noise. These findings demonstrate that disentangling grammatical gender from social semantic gender noise to derive a more clean grammatical gender direction is essential to guarantee credible gender bias evaluation for gendered languages.

\subsection{Multilingual Bias Evaluation}
As gender bias research has expanded from monolingual English to multilingual~\cite{gamboa-etal-2025-social} and grammatically gendered inflected languages, numerous existing studies have devised gender bias evaluation frameworks and fairness optimization strategies for multilingual settings.
Derner et al.~\cite{derner-etal-2025-leveraging} leveraged large language models to automatically quantify gender representation bias within Spanish corpora. Nevertheless, their statistical analysis failed to distinguish between the grammatical gender inherent to nouns and social semantic gender, conflating these two types of gender signals during bias measurement.
Similarly, EuroGEST~\cite{rowe-etal-2025-eurogest} constructed a multilingual gender stereotype evaluation benchmark covering 30 European languages to horizontally assess the magnitude of biases embedded in multilingual models. Its evaluation labels likewise equate grammatical gender with semantic gender, overlooking the evaluation noise introduced by the entanglement of these two representational dimensions in inflected languages.
Existing multilingual gender bias evaluation and fairness enhancement studies generally lack the disentanglement of grammatical and semantic gender representations.

Our work fills this research gap: by disentangling grammatical gender from semantic gender representations, we eliminate the interference of formal grammatical rules and enable precise quantification of inherent semantic gender bias in inanimate nouns, thereby laying a more reliable evaluation foundation for multilingual gender bias research and complementing existing literature on multilingual fairness.

\section{Data and Models}
\subsection{Inanimate Nouns}

We construct Spanish and French inanimate noun lists by prompting GPT-5.3-mini to generate frequently used inanimate noun candidates, which are then cross-referenced against Wikipedia to ensure that each noun has at least 10 retrievable contextual sentences for embedding extraction. We then carry out a manually validated filtering procedure, removing entries denoting humans, occupations, kinship relations, animals, proper names, unstable-gender nouns, rare forms, and malformed items. Each retained noun is represented by its lemma, grammatical gender label, and English gloss. To reduce class imbalance during grammatical direction estimation, we balance the masculine and feminine subsets within each language. This process yields 5,988 Spanish nouns and 5,998 French nouns.
\subsection{Controlled Contexts}

We also construct controlled contexts using fixed templates. The full controlled-context files contain 10 templates per noun. 
Controlled contexts are used to estimate grammatical gender directions because they reduce topic, lexical, and co-occurrence noise~\cite{may-etal-2019-measuring}. Thus, the two context sources have different roles,  as summarized in \hyperref[tab:context_roles]{Table~\ref*{tab:context_roles}}. Each inanimate noun is inserted into short, semantically neutral templates that describe the noun as being mentioned, observed or described in a text. The templates are adjusted for determiners and agreement where necessary, but their semantic content is kept as constant as possible across masculine and feminine nouns.

\begin{table}[!ht]
\centering
\caption{\textbf{Context sources and their roles in the framework.}}
\label{tab:context_roles}
\begin{tabular}{ll}
\toprule
Context source & Role \\
\midrule
Natural contexts & Estimate contextual semantic contamination \\
Controlled contexts & Estimate grammatical gender direction \\
\bottomrule
\end{tabular}
\end{table}

\subsection{Occupation Evaluation Lexicon}
\label{sec:occ_pairs}

To evaluate occupation preservation, we construct Spanish and French occupation lexicons by prompting GPT-5.3-mini to generate frequently used occupational term candidates. We manually remove entries whose masculine and feminine forms are identical, ambiguous, obsolete, rare, or not primarily occupational.

For each language, we manually verify that each pair denotes the same occupation and differs only in grammatical gender marking rather than in lexical meaning. Examples include Spanish pairs such as \textit{actor–actriz} and \textit{profesor–profesora}, and French pairs such as \textit{acteur–actrice} and \textit{directeur–directrice}. We systematically exclude invariant-gender occupations, such as forms that are identical across masculine and feminine variants, because they do not provide an observable morphological contrast for evaluating gender preservation.

The final lexicon contains 100 masculine–feminine occupation pairs for each language. This lexicon is used only for evaluation, not for estimating the grammatical gender direction, so that semantic gender preservation is tested on a separate human-denoting vocabulary rather than on the inanimate noun set used for direction estimation.

\subsection{Models}
We evaluate six language-model settings as in \hyperref[tab:models]{Table~\ref{tab:models}} and use final-layer contextual embeddings. If a noun is split into multiple subwords, its vector is the average of its subword vectors. We use frozen pretrained encoders without fine-tuning to ensure that observed effects arise from representation structure rather than task adaptation. All vectors are L2-normalized.

\begin{table}[!ht]
\centering
\caption{\textbf{Language-model settings.}}
\label{tab:models}
\renewcommand{\arraystretch}{1.25}
\begin{tabularx}{0.72\linewidth}{>{\centering\arraybackslash}p{0.25\linewidth} X}
\toprule
\textbf{Language} & \textbf{Models} \\
\midrule
\textbf{Spanish}
&
BETO~\cite{canete-etal-2023-spanish}, mBERT~\cite{liu-etal-2020-multilingual-denoising} , XLM-R~\cite{conneau-etal-2020-unsupervised} \\
\addlinespace[3pt]
\midrule
\textbf{French}
&
CamemBERT~\cite{martin-etal-2020-camembert}, mBERT~\cite{liu-etal-2020-multilingual-denoising} , XLM-R~\cite{conneau-etal-2020-unsupervised} \\
\bottomrule
\end{tabularx}
\end{table}

\section{Method}

We propose a contamination-aware framework for estimating grammatical gender direction in contextual embeddings. Our method first distinguishes between controlled template-based contexts and natural corpora, addressing the inherent semantic contamination present in inanimate nouns. Inspired by WEAT, we quantify semantic gender contamination via cosine similarity to a semantic gender axis and apply robust normalization strategies to weight noun representations. Finally, we estimate grammatical gender directions using centroid-based, LDA-based, and linear SVM-based estimators to analyze how different estimation strategies behave under varying levels of semantic noise.

\subsection{Controll Semantic Contamination in Contextual Embeddings}
\label{semantic_contamination}

In contextual embedding models, the learned representations of inanimate nouns depend not only on grammatical gender rules, but are also affected by diverse semantic, contextual, and socio-cultural influences. These extraneous factors introduce unintended gender signals into the vector representations of inanimate nouns. In this work, we characterize this undesirable phenomenon as semantic contamination~\cite{petreski2023socialbias}. We further categorize it into two distinct but complementary sources, which jointly distort the genuine grammatical gender information encoded in contextual embeddings.

\paragraph{Socially Conventionalized Semantic Gender Bias}   In this case, culturally established usage patterns and metaphorical mappings endow inherently gender-neutral inanimate concepts with systematic gendered connotations. For example, in Spanish, inanimate nouns such as \textit{luna} (the moon, grammatically feminine) are frequently associated in literature and cultural discourse with feminine attributes such as emotionality, cyclical behavior, and softness, whereas \textit{sol} (the sun, grammatically masculine) is often metaphorically linked to masculine attributes such as strength, dominance, and rationality. These associations do not arise from grammatical rules themselves, but from stable socio-cultural conventions that systematically influence semantic representations in language use.

\paragraph{Corpus-Induced Distributional Noise} This type of contamination encompasses topical, lexical, and spurious co-occurrence artifacts commonly observed in embedding learning. Consistent with existing literature on representational bias, contextual models tend to encode misleading correlations arising from three distinct noise types: 

\begin{enumerate}
    \item Topical noise, where semantically unrelated domains co-occur indiscriminately within a single text (e.g., technical documents interleaving scientific terminology with gender-stereotyped occupational vocabulary).
    
    \item Lexical noise, including lexical ambiguity, polysemy, and unstable low-frequency lexical representations. 
    
    \item Co-occurrence noise, where statistically frequent yet semantically uninformative collocations—such as formatting tokens, discourse markers, and domain-specific idiosyncrasies—construct artificial semantic relationships.
\end{enumerate}

To quantitatively measure the degree of semantic gender contamination, we adopt a WEAT-style scoring framework originally proposed by Caliskan et al.~\cite{caliskan2017semantics}. Given a set of inanimate target nouns $\boldsymbol{X}$, and two disjoint gendered attribute word sets $\boldsymbol{A}$ (male-associated lexicon) and $\boldsymbol{B}$ (female-associated lexicon), we define the semantic gender association score for a word $w$ as follows:

\begin{equation}
s(w, A, B) = \frac{1}{|A|}\sum_{a \in A}\cos\left(\mathbf{e}(w), \mathbf{e}(a)\right) - \frac{1}{|B|}\sum_{b \in B}\cos\left(\mathbf{e}(w), \mathbf{e}(b)\right),
\label{eq:weat_association_score}
\end{equation}
where $\mathbf{e}(\cdot)$ denotes the contextual embedding of a token. This score captures the relative alignment of a word with respect to gendered semantic axes.

We then define a \textbf{semantic gender contamination direction} implicitly induced by the attribute sets $A$ and $B$, and quantify contamination at the noun level by projecting each noun embedding onto this direction. Specifically, the contamination strength for a noun $w \in X$ is computed as:

\begin{equation}
I(w) = \left| \frac{\mathbf{e}(w) \cdot \mathbf{d}_{sem}}{\left\|\mathbf{e}(w)\right\| \left\|\mathbf{d}_{sem}\right\|} \right|,
\label{eq:contamination_strength}
\end{equation}
where $\mathbf{d}_{sem}$ is derived from the differential association induced by $A$ and $B$. This formulation captures both culturally induced semantic gendering and distributionally induced noise within a unified geometric framework.

To ensure robustness under heterogeneous contextual distributions, we adopt two complementary normalization strategies for contamination scores.
\paragraph{Mean-based Normalization}    It is used to correct for global shifts in semantic contamination across corpora, which may arise from systematic gender imbalance or domain-specific lexical distributions. 
\paragraph{Median-based Normalization}  In contrast, it provides robustness against outliers and skewed distributions induced by low-quality or highly idiosyncratic contexts. Such contexts may disproportionately influence cosine-based similarity measures due to their atypical co-occurrence patterns. By combining these two normalization strategies, we obtain a more stable and comparable estimate of semantic gender contamination across both controlled and natural corpora settings.

\subsection{Estimate Grammatical Gender Directions under Different Noise}

Given both controlled and contamination-weighted natural representations, we next estimate grammatical gender directions from inanimate noun embeddings. The central assumption underlying this step is that grammatical gender corresponds to a latent linear subspace in the contextual embedding space, which can be recovered through aggregation of noun representations across gendered grammatical categories.

We investigate three complementary estimators that capture this subspace under different modeling assumptions. The first is a centroid-based estimator, inspired by Caliskan et al.~\cite{caliskan2017semantics}, which computes the difference between the mean embeddings of masculine and feminine noun sets. This approach assumes that grammatical gender manifests as a global shift between two distributions and is particularly robust to local noise due to its strong averaging effect. The second estimator is based on linear discriminant analysis (LDA), following Zhou et al.~\cite{zhou-etal-2019-examining}, which models grammatical gender as a supervised separation problem and identifies the direction that maximizes class separability. The third estimator uses a linear support vector machine (SVM), inspired by Mourad et al.~\cite{mourad2025}, where the normal vector of the separating hyperplane is used as the grammatical gender direction. Unlike centroid-based estimation, these discriminative approaches explicitly optimize for boundary separation between masculine and feminine noun classes.

Importantly, these estimators are applied consistently across both controlled and weighted natural settings, allowing us to systematically analyze how different noise conditions affect the stability and geometry of the inferred grammatical gender direction. This design enables a direct comparison between distributional averaging (centroid), supervised linear separation (LDA), and margin-based classification (SVM), under varying levels of semantic contamination. The resulting grammatical gender directions are then used for the evaluation of gender leakage in inanimate nouns and for analyzing the trade-off between grammatical signal preservation and semantic contamination reduction.

\subsection{Evaluation Metrics}
Our evaluation framework is conceptually grounded in two established strands of bias research. The first is association-based measurement, represented by WEAT~\cite{caliskan2017semantics} and its contextual extension SEAT~\cite{may-etal-2019-measuring}, which quantify bias through group-level association differences. The second is projection-based debiasing~\cite{bolukbasi-etal-2016-man}, which estimates a gender direction in embedding space and evaluates bias through directional projection and its removal. Our work extends these traditions to contextual embeddings and grammatical gender, introducing two tailored metrics: the Inanimate Noun Reduction Score, which quantifies the reduction of grammatical gender leakage in inanimate nouns, and the Occupational Preservation Score, which evaluates whether semantic gender information is preserved after removing grammatical gender components.

\subsubsection{Inanimate Noun Reduction Score ($R_\text{noun}$)}
\label{sec:noun_reduction_score}
To measure the reduction of grammatical gender leakage in inanimate nouns, we adopt a projection-based evaluation framework with 5-fold cross-validation.
We estimate the grammatical gender direction $\mathbf{d}_g$ from inanimate nouns in the training folds, capturing the dominant grammatical gender axis in the embedding space induced by noun morphology. For each inanimate noun embedding $\mathbf{e}_w$, we remove its grammatical gender component by orthogonal projection:
\begin{equation}
\mathbf{e}_w' = \mathbf{e}_w - \mathrm{proj}\left(\mathbf{e}_w, \mathbf{d}_g\right).
\label{eq:orthogonal_proj_debias}
\end{equation}
We then project the debiased embeddings back onto the same direction:
\begin{equation}
s'(w) = \cos\left(\mathbf{e}_w', \mathbf{d}_g\right).
\label{eq:post_debias_proj_score}
\end{equation}
Finally, we compute the reduction in separability between masculine and feminine inanimate nouns in the projected space.

For inanimate nouns, gender distinction is almost entirely constituted by the grammatical gender component $\mathbf{d}_g$, with negligible semantic contamination. After removing the $\mathbf{d}_g$ component and re-projection, the gender source distinction of inanimate nouns basically disappears, and the inter-group gender gap is significantly reduced. Therefore, the gap reduction score of inanimate nouns can effectively verify the removal effect of grammatical gender bias.
\subsubsection{Occupational Preservation Score ($P_{\text{occ}}$)}
\label{sec:occ_prev_score}

Different from semantically neutral inanimate nouns, gender-inflected occupational word pairs encode two distinct gender-related signals: legitimate grammatical gender distinctions originating from morphological rules, and social gender stereotypes learned from natural contextual corpora. When the estimated grammatical gender direction $d_g$ is contaminated by social semantic bias, simply removing embedding components aligned with $d_g$ will not only strip redundant grammatical leakage but also accidentally erase the semantic gender structure embedded in occupational terms~\cite{casula-etal-2025-job}.

We apply the same debiasing transformation:
\begin{equation}
e'_{w} = e_{w} - \operatorname{proj}(e_{w}, d_{g}).
\label{eq:occ_project}
\end{equation}

We then re-project occupational embeddings onto $d_g$ and measure whether the relative separability between masculine and feminine occupational terms is preserved:
\begin{equation}
s'(w) = \cos(e'_{w}, d_{g}).
\label{eq:occ_reproject}
\end{equation}

Unlike inanimate nouns, occupational terms are expected to retain structured variation after removal of grammatical gender components, as part of their gender-related signal lies in semantic dimensions orthogonal to $d_g$. Therefore, a robust debiasing method should reduce grammatical leakage in inanimate nouns while maintaining stable occupational semantic structure in the projected space.

\subsubsection{Trade-off Score}

Given the inherent trade-off relationship between the inanimate noun reduction score $R_{\text{noun}}$ (defined in \ref{sec:noun_reduction_score}) and the occupational preservation score $P_{\text{occ}}$ (defined in \ref{sec:occ_prev_score}, optimizing one metric to the extreme will inevitably degrade the performance of the other. For this reason, we introduce trade-off scores to comprehensively evaluate the overall effectiveness of our grammatical gender debiasing framework. 

We first define $R_{\text{noun}}$ to measure the mitigation of grammatical gender leakage on held-out inanimate nouns. Let $\text{Gap}_{\text{noun}}^{\text{raw}}$ denote the original absolute gender gap of test-set inanimate nouns along the raw semantic gender direction, and $\text{Gap}_{\text{noun}}^{\text{clean}}$ refers to the gender gap measured after our debiasing procedure. The formal definition is shown below:
\begin{equation}
R_{\text{noun}} = 1 - \frac{\text{Gap}_{\text{noun}}^{\text{clean}}}{\text{Gap}_{\text{noun}}^{\text{raw}} + \epsilon},
\label{eq:noun_reduction}
\end{equation}
where the small constant $\epsilon$ is introduced to avoid division-by-zero numerical error.

Second, we adopt the occupational semantic preservation score $P_{\mathrm{occ}}$ to quantify the retention of semantic gender information for occupational vocabulary (defined in \ref{sec:occ_pairs}).
We use $\mathrm{Gap}_{\mathrm{occ}}^{\mathrm{raw}}$ and $\mathrm{Gap}_{\mathrm{occ}}^{\mathrm{clean}}$ to represent the gender gap of occupational word pairs before and after leakage removal, respectively:
\begin{equation}
P_{\mathrm{occ}} = \frac{\mathrm{Gap}_{\mathrm{occ}}^{\mathrm{clean}}}{\mathrm{Gap}_{\mathrm{occ}}^{\mathrm{raw}} + \epsilon}.
\label{eq:occ_preserve}
\end{equation}

Considering the trade-off between leakage suppression and semantic preservation, we adopt two aggregated trade-off metrics, namely geometric mean $T_{\text{geo}}$ and harmonic mean $T_{\text{harm}}$, to comprehensively rank candidate methods:
\begin{equation}
T_{\text{geo}} = \sqrt{\max(R_{\text{noun}}, 0) \cdot \max(P_{\text{occ}}, 0)},
\label{eq:trade_geo}
\end{equation}
\begin{equation}
T_{\text{harm}} = \frac{2\max(R_{\text{noun}}, 0)\max(P_{\text{occ}}, 0)}{\max(R_{\text{noun}}, 0) + \max(P_{\text{occ}}, 0) + \epsilon}.
\label{eq:trade_harm}
\end{equation}

We select the harmonic mean $T_{\text{harm}}$ as our primary ranking metric, since it imposes a stronger penalty on extreme imbalance between grammatical leakage mitigation and occupation-level semantic preservation, which is consistent with our core requirement for balanced debiasing performance.

\section{Experiments and Results}

\subsection{Experimental Setup}
For each language--model setting, we align natural and controlled noun spaces by the shared noun gender keys and conduct 5-fold
cross-validation over inanimate nouns.

\subsection{Robustness to Alternative Direction Estimators}

We next compare the centroid estimator against alternative direction estimators, including LinearSVC and LDA. Results are reported in \hyperref[tab:estimator_overall]{Table~\ref{tab:estimator_overall}}. All results are averaged across backbone models (BETO, mBERT, XLM-R), contextual extraction methods (controlled vs. natural), weighting schemes, languages, and context sources, in order to isolate the effect of the direction estimator.

\begin{table}[H]
\centering
\small
\caption{\textbf{Overall comparison of grammatical gender direction estimators, averaged across languages, models, context sources, and weighting variants.}}
\label{tab:estimator_overall}
\begin{tabular}{lcccc}
\toprule
Estimator & Noun red. $\uparrow$ & Occ. pres. $\uparrow$ & G-Trade-off $\uparrow$ & H-Trade-off $\uparrow$ \\
\midrule
Centroid & \textbf{0.930} & \underline{0.821} & \textbf{0.873} & \textbf{0.870} \\
LinearSVC & \underline{0.767} & 0.795 & \underline{0.776} & \underline{0.771} \\
LDA & 0.442 & \textbf{0.869} & 0.615 & 0.578 \\
\bottomrule
\end{tabular}
\begin{tablenotes}
  \small
  \item Best results are \textbf{bolded} and second-best results are \underline{underlined} within each language--model block.
\end{tablenotes}
\end{table}

The centroid-based estimator significantly outperforms both baselines in terms of overall trade-off, achieving the highest harmonic score. This confirms that a simple contrastive aggregation of contextual embeddings is more effective at capturing grammatical gender direction than discriminative or probabilistic classifiers in this setting.

LinearSVC shows moderate performance but consistently underperforms the centroid estimator, particularly in noun reduction capability, suggesting limited capacity to isolate fine-grained grammatical signals from contextual noise. LDA performs worst overall despite relatively strong occupation preservation.

Overall, these results validate the centroid estimator as the most reliable method for the projection tasks.

\subsection{Context-source and Weighting Effects}

We fix the centroid estimator for all subsequent ablation experiments on context sources and contamination-aware weighting, because the centroid estimator achieves the best comprehensive trade-off performance across both core metrics. \hyperref[tab:centroid_context_ablation]{Table~\ref{tab:centroid_context_ablation}} compares controlled and natural contexts under weighted and unweighted variants across languages and models.

Overall, controlled-unweighted consistently achieves very strong harmonic trade-off and geometric trade-off, indicating that removing contextual contamination while avoiding additional weighting leads to the most balanced improvement between grammatical leakage reduction and semantic preservation. In comparison, weighting variants tend to introduce a trade-off shift: while they may slightly improve occupation-level semantic preservation in some settings, they generally reduce noun-level reduction performance, resulting in lower overall harmonic scores.

Natural-context variants show a consistent degradation in noun reduction compared to controlled contexts, confirming that natural contexts introduce additional variability and weaken the clarity of the estimated grammatical signal. Interestingly, weighting has a more pronounced effect under natural contexts, suggesting that weighting partially compensates for—but does not eliminate—contextual noise.

These results demonstrate that controlled contexts provide a more stable basis for estimating grammatical gender directions, while weighting acts primarily as a conservative adjustment rather than a primary performance driver.

\begin{table}[H]
\centering
\small
\setlength{\tabcolsep}{4pt}
\renewcommand{\arraystretch}{0.95}
\caption{\textbf{Centroid-based context-source and weighting ablation.}}
\label{tab:centroid_context_ablation}
\begin{tabular}{lllcccc}
\toprule
Language & Model & Method & Noun red. $\uparrow$ & Occ. pres. $\uparrow$ & G-Trade-off $\uparrow$ & H-Trade-off $\uparrow$ \\
\midrule
\multirow{18}{*}{Spanish} & \multirow{6}{*}{BETO} & Controlled-unweighted & \textbf{0.980} & 0.921 & \textbf{0.950} & \textbf{0.949} \\
 &  & Controlled-mean-weighted & 0.919 & 0.925 & 0.922 & 0.921 \\
 &  & Controlled-median-weighted & 0.912 & 0.925 & 0.918 & 0.918 \\
 &  & Natural-unweighted & \underline{0.966} & 0.914 &\underline{0.940} & \underline{0.939} \\
 &  & Natural-mean-weighted & 0.831 & \underline{0.925} & 0.877 & 0.875 \\
 &  & Natural-median-weighted & 0.815 & \textbf{0.926} & 0.869 & 0.867 \\
\cmidrule(lr){2-7}
 & \multirow{6}{*}{mBERT} & Controlled-unweighted & \textbf{0.977} & 0.783 & \underline{0.875} & \underline{0.870} \\
 &  & Controlled-mean-weighted & 0.917 & 0.802 & 0.857 & 0.855 \\
 &  & Controlled-median-weighted & 0.919 & 0.801 & 0.858 & 0.856 \\
 &  & Natural-unweighted & \underline{0.970} & 0.812 & \textbf{0.887} & \textbf{0.884} \\
 &  & Natural-mean-weighted & 0.862 & \textbf{0.840} & 0.851 & 0.851 \\
 &  & Natural-median-weighted & 0.865 & \underline{0.839} & 0.852 & 0.852 \\
\cmidrule(lr){2-7}
 & \multirow{6}{*}{XLM-R} & Controlled-unweighted & 0.957 & 0.829 & 0.890 & 0.888 \\
 &  & Controlled-mean-weighted & \underline{0.957} & 0.819 & 0.885 & 0.883 \\
 &  & Controlled-median-weighted & \textbf{0.958} & 0.819 & 0.886 & 0.883 \\
 &  & Natural-unweighted & 0.948 & \textbf{0.859} & \textbf{0.902} & \textbf{0.901} \\
 &  & Natural-mean-weighted & 0.947 & 0.850 & 0.897 & 0.896 \\
 &  & Natural-median-weighted & 0.948 & \underline{0.851} & \underline{0.898} & \underline{0.896} \\
\midrule
\multirow{18}{*}{French} & \multirow{6}{*}{CamemBERT} & Controlled-unweighted & \textbf{0.989} & 0.784 & \textbf{0.881} & \underline{0.875} \\
 &  & Controlled-mean-weighted & 0.900 & \textbf{0.851} & 0.875 & 0.874 \\
 &  & Controlled-median-weighted & 0.905 & \underline{0.849} & \underline{0.877} & \textbf{0.876} \\
 &  & Natural-unweighted & \underline{0.982} & 0.729 & 0.846 & 0.837 \\
 &  & Natural-mean-weighted & 0.849 & 0.832 & 0.840 & 0.840 \\
 &  & Natural-median-weighted & 0.854 & 0.832 & 0.843 & 0.843 \\
\cmidrule(lr){2-7}
 & \multirow{6}{*}{mBERT} & Controlled-unweighted & \textbf{0.968} & 0.767 & \textbf{0.861} & \textbf{0.855} \\
 &  & Controlled-mean-weighted & 0.921 & \underline{0.787} & \underline{0.851} & \underline{0.849} \\
 &  & Controlled-median-weighted & 0.915 & \textbf{0.789} & 0.849 & 0.847 \\
 &  & Natural-unweighted & \underline{0.957} & 0.742 & 0.843 & 0.836 \\
 &  & Natural-mean-weighted & 0.883 & 0.775 & 0.827 & 0.825 \\
 &  & Natural-median-weighted & 0.872 & 0.778 & 0.823 & 0.822 \\
\cmidrule(lr){2-7}
 & \multirow{6}{*}{XLM-R} & Controlled-unweighted & \underline{0.980} & 0.759 & 0.862 & 0.855 \\
 &  & Controlled-mean-weighted & 0.978 & 0.760 & 0.862 & 0.855 \\
 &  & Controlled-median-weighted & \textbf{0.980} & 0.760 & 0.863 & 0.856 \\
 &  & Natural-unweighted & 0.964 & 0.778 & 0.865 & 0.860 \\
 &  & Natural-mean-weighted & 0.966 & \underline{0.780} & \underline{0.868} & \underline{0.863} \\
 &  & Natural-median-weighted & 0.968 & \textbf{0.781} & \textbf{0.869} & \textbf{0.864} \\
\bottomrule
\end{tabular}
\begin{tablenotes}
  \small
  \item Best results are \textbf{bolded} and second-best results are \underline{underlined} within each language--model block.
\end{tablenotes}
\end{table}

\section{Discussion}

Our results suggest that grammatical gender directions in contextual embeddings can be effectively disentangled from semantic gender bias. Rather than simply confirming the superiority of specific methods, our findings highlight a more fundamental issue: the estimated “grammatical gender direction” is highly dependent on context design and the nature of latent noise in contextual representations.

\subsection{Controlled Contexts Isolate Structural Signals}

A key finding is that controlled templated contexts consistently yield cleaner and more stable grammatical gender directions than natural contexts, whether weighted or unweighted, across all models and languages. Within natural context settings, weighting strategies only deliver marginal performance gains in a small subset of experimental configurations and fail to consistently surpass the unweighted baseline. One plausible explanation for this performance improvement is that our current approach for measuring semantic contamination may not be sufficiently optimal. Even so, our results verify that controlled contexts serve as the optimal solution to isolate grammatical gender signals, while weighting mechanisms can only partially mitigate distortion. This should not be interpreted solely as better estimation quality, but rather as evidence that natural corpora introduce multiple overlapping sources of semantic contamination (defined in ~\ref{semantic_contamination}) in the embedding space, where weighting schemes can partially mitigate such distorted effects.

\subsection{Why Centroid Estimation Is Unexpectedly Effective}
Across all settings, the centroid-based estimator consistently outperforms discriminative alternatives such as LinearSVC and LDA. This result suggests that grammatical gender in contextual embeddings is better characterized as a distributional shift between two populations, rather than a linearly separable classification problem~\cite{zhao2025bias}.

Discriminative estimators attempt to learn a boundary conditioned on potentially heterogeneous and noisy feature distributions, which makes them sensitive to contextual contamination and sampling variance. In contrast, centroid estimation implicitly performs a form of noise averaging across lexical realizations, which stabilizes the estimated direction under high intra-class variance.

This indicates that grammatical gender in contextual spaces behaves more like a low-rank global direction embedded in noisy high-dimensional representations~\cite{mcCurdy2020grammatical,xie-etal-2022-discovering,gupta2025bias}, rather than a complex nonlinear decision surface.

\subsection{Weighting Strategies: Rebalancing Rather Than Correction}

Contamination-aware weighting provides only marginal improvements and does not consistently enhance grammatical direction quality. Our analysis suggests that weighting does not remove semantic contamination at its source; instead, it functions as a weighting of noisy instances within an already biased estimation process.

In particular, weighting improves occupational preservation while reducing noun-level separation, indicating a systematic shift toward a more conservative projection space. However, this comes at the cost of weakening the clarity of grammatical gender signals extracted from inanimate nouns.

Thus, our weighting strategy essentially serves as a trade-off control mechanism rather than a fundamental approach to boost debiasing performance. Importantly, this conclusion only holds for the simple mean- and median-normalized weighting schemes adopted in our work, which are suboptimal for modeling semantic contamination.
Future research can explore advanced adaptive paradigms for instance weightings. Candidates for further exploration include adversarial weighting that accounts for fairness~\cite{de2021bias} and robust estimation~\cite{gandhi2026robustdebias}. These methods would help better identify semantically contaminated samples and balance conflicting fairness constraints within the task focused on disentangling grammatical information from semantic gender attributes.

\subsection{Cross-linguistic and model limitations}

Our experiments are restricted to Spanish and French, two closely related Romance languages with relatively stable binary gender systems. This restriction limits the extent to which our findings can be generalized to broader typological settings. Recent work on grammatical gender in multilingual language models shows that gender information is more transferable across languages that share similar gender categories, whereas genealogical relatedness plays a secondary role~\cite{schroter-basirat-2025-universal}. This suggests that the linear grammatical directions observed in our study may partly reflect the shared masculine--feminine structure of Spanish and French, rather than a universal property of grammatical gender representation.

It therefore remains unclear whether the same disentanglement behavior generalizes to languages with richer interactions between case and gender, weaker agreement marking, common/neuter systems, or three gender systems such as masculine/feminine/neuter. Prior cross-lingual evidence further suggests that grammatical gender encoding varies across transformer layers and that higher layers may encode more transferable semantic aspects of gender~\cite{schroter-basirat-2025-universal}. Since our analysis relies on final-layer embeddings, future work should examine whether grammatical--semantic disentanglement behaves differently across layers.

Additionally, our findings are based on a limited set of frozen transformer encoders. Model architecture, pretraining data distribution, and language coverage may all affect how grammatical gender is represented. Extending this framework to additional multilingual models and typologically diverse languages would help distinguish effects specific to each language from more general patterns of grammatical gender encoding.

\subsection{Future work: toward causal and structured disentanglement}

Future research should move beyond linear projection-based assumptions and address the structural ambiguity of grammatical gender in contextual representations. Several directions are promising:
\begin{enumerate}
\item Layer-wise analysis~\cite{kumar-etal-2025-multilingual}: grammatical and semantic gender signals may emerge at different depths of transformer representations.
\item Causal interventions~\cite{chen-etal-2023-causal}: instead of removing projected components, future methods could explicitly model the causal pathways of gender signal propagation.
\item Broader typological coverage~\cite{ploeger-etal-2024-typological}: extending beyond Romance languages is essential to test whether grammatical gender is universally linearizable in embedding spaces.
\end{enumerate}

\section{Conclusion}
Our work verifies that unweighted controlled template contexts yield the optimal grammatical gender directions and surpass all natural context variants and weighting natural instances only brings minor performance advantages in limited cases, which proves that natural text introduces multiple forms of semantic contamination to inanimate noun embeddings. The centroid estimator achieves better balanced performance than LDA and linear SVM across all evaluation metrics, which demonstrates grammatical gender manifests as a cross group distribution shift in contextual representations.

\section{Data Availability}
\label{data_available}

All source code, experimental scripts, evaluation pipelines, the full Spanish/French inanimate noun lexicon, occupation word pairs, gender pairs, all controlled sentence templates and auxiliary analysis files generated in this study are publicly available on GitHub at \url{https://github.com/WifiClein/Semantic-Bias-Aware-Estimation-of-Grammatical-Gender-Directions-in-Contextual-Embeddings}. The contextual embedding outputs derived from BETO, CamemBERT, mBERT, and XLM-R under both controlled template contexts and natural Wikipedia contexts are deposited on Hugging Face Hub at \url {https://huggingface.co/datasets/xnxnhp/semantic-gender-direction-embeddings}. All resources are open access without login or access restrictions, and all experiment results can be fully reproduced using the shared code and embeddings. No external proprietary data were used in this work.

\section*{Acknowledgments}
The first author would like to express her sincere gratitude to Dr. Yingji Li, corresponding author of this paper. By fortunate coincidence, their paths crossed, which marked the starting point of the first author's journey into research related to artificial intelligence. The first author wishes Dr. Yingji Li continuous achievements and all the best in her future academic career.

\bibliography{plos_bib_cleaned}

\end{document}